\documentclass[conference]{IEEEtran}
\usepackage{cite}
\usepackage{amsmath,amssymb,amsfonts}
\usepackage{graphicx}
\usepackage{booktabs}
\usepackage{multirow}
\usepackage{xcolor}
\usepackage[hidelinks]{hyperref}
\usepackage{url}
\hypersetup{
  pdfauthor={Zihao Zheng, Junyi Yao, Baichuan Li, and Jiayu Long},
  pdftitle={Auditing Escalation Reliability in Healthcare LLM Agents: A Perturbation-Controlled Benchmark of Workflow-State Failures},
  pdfsubject={Healthcare LLM agent evaluation and escalation reliability}
}

\newcommand{\actA}{\textsc{Answer}}
\newcommand{\actK}{\textsc{Ask\_Clarification}}
\newcommand{\actB}{\textsc{Abstain}}
\newcommand{\actE}{\textsc{Escalate}}

\begin{document}

\title{Auditing Escalation Reliability in Healthcare LLM Agents:\\
A Perturbation-Controlled Benchmark of\\
Workflow-State Failures}

\author{\IEEEauthorblockN{Zihao Zheng\IEEEauthorrefmark{1},
Junyi Yao\IEEEauthorrefmark{1},
Baichuan Li\IEEEauthorrefmark{2}, and
Jiayu Long\IEEEauthorrefmark{1}}
\IEEEauthorblockA{\IEEEauthorrefmark{1}Washington University in St. Louis, USA}
\IEEEauthorblockA{\IEEEauthorrefmark{2}Southern Methodist University, USA}}

\maketitle

\begin{abstract}
Trustworthy biomedical AI requires reliable action selection when evidence
is missing, stale, conflicting, or late, not only accurate answers on static
inputs. We instantiate an episode-level evaluation protocol as 400 synthetic
healthcare workflow episodes: 40 hand-authored semantic seeds across four
task classes, two surface parameterizations, and five controlled variants,
scored on \actA{}/\actK{}/\actB{}/\actE{} under a pre-declared 5:1
missed-versus-unnecessary escalation cost. Reference actions are
single-author study-policy labels, not clinician-adjudicated ground truth.
Across five open-weight configurations (1.7B--9.7B), clean decision accuracy
reaches 0.97, yet uncertainty-prompted models miss 32--97\% of required
escalations. The stronger 9.7B configuration narrows but does not remove the
gap, falling from 0.97 clean accuracy to 0.78 under conflict. Verbalized
confidence is inconsistent (ECE 0.10--0.54; failure AUROC 0.48--0.77), and a
development-tuned confidence-only policy yields no utility gain. A
transparent workflow-aware controller improves 5:1 utility for every model
and reduces missed escalations to 0.12--0.24, at specificity 0.83--0.93.
On the single-action subset, the 9.7B model's utility gain disappears.
However, a 30-concept lexical stress test reduces each rule detector's recall
from 1.00 on trigger phrases to 0.00 on unseen paraphrases, limiting the
controller claim to this benchmark. Seed-clustered statistics, ablations,
cross-task transfer, raw outputs, and manifests are packaged for audit. The
results motivate explicit workflow-state evaluation while showing that
apparently effective routing controls need their own generalization tests.
\end{abstract}


\section{Introduction}
Trustworthy biomedical AI requires more than accurate predictions on static
inputs: systems must remain reliable as evidence changes, conflicts, or
arrives late. Healthcare work is sequential and stateful. A clinical assistant embedded in
documentation, messaging, or triage workflows receives information out of
order, encounters contradictions between sources, and sees risk signals
appear after a task has already been posed. In such settings the safety-critical
competence is not fluent generation but \emph{selective decision-making}:
knowing when to complete the task, when to request a missing fact, when to
decline, and when to hand off to a human immediately.

Prior work introduced an episode-level evaluation protocol for healthcare
NLP agents~\cite{yao2026workflow}. It separates model-level competence from
agent-level behavior in simulated workflow episodes and specifies a five-field
episode schema, four task templates, and a cost-sensitive treatment of missed
versus unnecessary escalation. That study released no dataset and reported no
experiments; its stated next step was a governed pilot with episodes,
adjudication decisions, score distributions, and failure analyses.

This paper provides an experimental instantiation, but not the completed
clinical validation envisioned by that protocol. We construct a perturbation-controlled
benchmark of 400 multi-turn episodes following the published schema, run
five open-weight LLM configurations under three control conditions, and answer three
empirical questions the protocol only made askable:

\begin{itemize}
\item \textbf{RQ1:} Do controlled workflow perturbations---missing, stale,
conflicting, and late-arriving information---degrade agents' decision
reliability even where clean-condition performance is high?
\item \textbf{RQ2:} Does model-level verbalized confidence predict decision
failures under workflow shift, i.e., can it serve as an agent-level control
signal?
\item \textbf{RQ3:} Can a lightweight, training-free selective controller
that combines model uncertainty with workflow-state indicators improve the
cost-sensitive tradeoff between missed and unnecessary escalation?
\end{itemize}

\noindent Our contributions are:
\begin{enumerate}
\item A 400-episode benchmark instantiating the workflow-aware protocol
of~\cite{yao2026workflow}: 40 hand-authored healthcare-workflow seeds $\times$ 2
surface parameterizations $\times$ 5 variants (clean/missing/stale/conflict/
late-risk) across four task classes, with four allowed actions, per-variant
reference decisions, and a paired clean--perturbed design.
\item An empirical characterization of an \emph{escalation deficit} in the
evaluated open-weight configurations: high clean-condition accuracy
coexisting with frequent missed escalations, model-specific failure signatures
(answering through conflicts vs.\ hedging instead of escalating), and the
finding that stale-but-salient summaries are handled far better than
contradictions.
\item A direct test of whether verbalized confidence transfers from model
level to agent-level escalation control in multi-turn episodes, using
calibration, failure discrimination, and confidence-only deferral baselines.
\item An audit of a transparent workflow-aware controller: in-distribution
utility, feature ablations, cross-task threshold transfer, and a diagnostic
lexical OOV stress set that exposes where simple detectors cease to be
reliable; plus the finding that prompt-level uncertainty instructions can
\emph{backfire}.
\end{enumerate}

The benchmark measures decision behavior in simulated episodes. Consistent
with the boundary discipline of~\cite{yao2026workflow}, it does not
establish clinical outcomes, deployment readiness, or institutional
generalization.

\section{Related Work}

\subsection{Abstention and escalation in medical LLMs}
SOSBench~\cite{jiang2026sosbench} evaluates scientific misuse, including
medicine and pharmacology. Medical benchmarks cover perturbed multiple-choice
abstention~\cite{machcha2026abstain}, one-shot triage
(MedSafe-Dx~\cite{vanoyen2026medsafe}), and diagnostic uncertainty
training~\cite{zhou2025confidx}. Multi-turn work studies clarification under
incomplete information (MediQ~\cite{li2024mediq}), evidence-sufficiency-based
abstention (KnowGuard~\cite{dang2026knowguard}), and unsafe agreement under
patient pressure (MedPRESS~\cite{joy2026medpress}).
AbstentionBench~\cite{kirichenko2025abstention} finds poor abstention on
general unanswerable queries; Wen et al.~\cite{wen2025limits} survey the
field. These studies do not jointly evaluate human escalation and controlled
changes in evolving workflow state.

\subsection{Multi-turn and agentic clinical evaluation}
MedAgentBench~\cite{jiang2025medagentbench} uses a live FHIR-compliant EHR;
HealthBench~\cite{arora2025healthbench} grades 5{,}000 health conversations
with physician rubrics including emergency referral.
CRAFT-MD~\cite{johri2025craftmd} and AgentClinic~\cite{schmidgall2026agentclinic}
show degradation during conversational information gathering.
AMIE~\cite{tu2025amie} evaluates diagnostic dialogue; its longitudinal
extension studies three-visit disease management~\cite{lievin2026management}.
Sequential diagnosis~\cite{nori2025sequential} includes information costs.
PatientAgentBench~\cite{vatanparvar2026patientagent} tests gather/act/escalate
decisions, but lacks paired perturbations, an \actB{} action, and
cost-sensitive scoring. MedDistractQA~\cite{vishwanath2025distract}
tests sensitivity to distracting clinical context. EHRNoteQA~\cite{kweon2024ehrnoteqa}
evaluates question answering grounded in clinical discharge summaries.
Our benchmark combines multi-turn episodes, paired workflow perturbations,
and four-way actions with cost-weighted escalation scoring.

\subsection{Selective prediction, calibration, and deferral}
We use established selective-risk analysis~\cite{geifman2017selective,kamath2020selective},
verbalized-confidence elicitation~\cite{tian2023just,xiong2024llms},
generation uncertainty~\cite{kuhn2023semantic,farquhar2024detecting,lin2024generating},
cost-sensitive deferral~\cite{madras2018predict,mozannar2020consistent},
uncertainty-routed cascades~\cite{gupta2024cascades,chen2023frugalgpt},
clinical deferral~\cite{strong2025guided}, and conformal LM
guarantees~\cite{mohri2024conformal,quach2024conformal} as measurement and
baseline context. Our contribution is the workflow-conditioned setting,
not new selective-prediction theory.
RAGScope~\cite{liu2026ragscope} studies low-cost evidence gates for
RAG review routing and exposes calibration failures under source shift;
we instead evaluate four-way actions under changing healthcare workflow state.

\section{Benchmark}

\subsection{Episode schema and task classes}
Each episode follows the five-field schema of~\cite{yao2026workflow}: an
initial context packet, a state-changing event, an allowed action space, a
reference decision profile, and an author-provided rationale. Episodes are
rendered as a transcript with typed blocks (setting, earlier record, new
information, request, and---where applicable---a message arriving after the
request). The agent must output exactly one action from
$\{$\actA{}, \actK{}, \actB{}, \actE{}$\}$ plus a confidence in $[0,1]$ and a
short reason, as a schema-enforced JSON object. Fixing the decision space
keeps the primary metric judge-free: an action either matches the
pre-specified acceptable set or it does not.

The four task classes follow the protocol's templates, with class-specific
action semantics: \emph{documentation update} (produce the note update vs.\
flag to a clinician), \emph{evidence retrieval} (answer strictly from a
supplied evidence packet vs.\ abstain on insufficiency or escalate on an
in-progress safety event), \emph{patient messaging} (draft a portal reply
vs.\ route to urgent care), and \emph{triage handoff} (route a nurse-line
call). Ten seed scenarios were hand-authored per class---for example, an
anticoagulant dose change after a high INR, a macrolide--statin interaction
check, a post-operative wound message, and an exertional chest-pain call.

\subsection{Controlled variants}
Every seed defines content hooks from which five variants are assembled
deterministically (no LLM generation), yielding paired episodes that differ
only in the manipulated axis:
\begin{itemize}
\item \textbf{Clean}: complete, unambiguous information;
\item \textbf{Missing}: a decision-critical fact is deleted;
\item \textbf{Stale}: an outdated auto-generated summary restating a
superseded fact is placed saliently after the update;
\item \textbf{Conflict}: a contradiction between sources, or patient
minimization of objectively concerning findings, is appended;
\item \textbf{Late-risk}: a red-flag cue arrives only \emph{after} the
request has been posed.
\end{itemize}
Reference actions are defined per variant, because the target decision
changes with the perturbation: a benign refill request is \actA{} when
clean, \actK{} when the dose is unstated, and \actE{} when thyrotoxic
symptoms arrive late. Multiple actions are acceptable in 62/400 episodes
(15.5\%): 16/80 development and 46/320 test episodes (14.4\%).
Decision accuracy scores membership in that set. Escalation metrics and
utility instead use the single reference action (\actE{} vs.\ other):
18 test episodes permit optional \actE{}, while 20 \actE{}-reference
episodes also accept a non-escalation action. Thus an acceptable action can
still be a binary false positive or false negative. These metrics answer
different policy-relative questions, not clinical correctness. The
reference-policy manual records all sets and labels; below we also exclude
all multi-action test episodes as a sensitivity analysis.

\subsection{Statistics, splits, and cost model}
The benchmark contains 40 seeds $\times$ 2 surface
parameterizations $\times$ 5 variants $=400$ episodes, 100 per task class;
130 episodes (32.5\%) have \actE{} as the reference action. Two seeds per
class (80 episodes) form a development split used exclusively for controller
tuning; all reported metrics use the 320 test episodes (100
escalation-required). Following the protocol's cost-sensitive utility
$U = u_{TP}TP + u_{TN}TN - c_{FN}FN - c_{FP}FP$ over escalation decisions,
we pre-declared $u_{TP}=u_{TN}=1$, $c_{FP}=1$, and $c_{FN}=5$ (primary),
with $c_{FN}=10$ as a robustness check, and report $U$ normalized per
episode. The 5:1 choice encodes greater aversion to a missed handoff;
10:1 stresses that assumption. Neither ratio estimates clinical harm or
measured staffing costs. A post-review analysis evaluates ratios
1:1, 2:1, 3:1, 5:1, 10:1, and 20:1 with the original policy fixed;
the artifact also repeats development-only tuning at each ratio.
All episodes are fully synthetic. Because two surface realizations
and five variants share one semantic seed,
statistical uncertainty is estimated by resampling the 32 test seeds, not
320 episodes.

\section{Methods}

\subsection{Generation conditions}
\textbf{Vanilla} presents the action space with neutral definitions.
\textbf{Uncertainty-prompted} adds an explicit checklist (check for missing,
stale, and conflicting information and for risk cues including
late-arriving ones), states that missed escalations are more harmful than
unnecessary ones, and instructs honest, calibrated confidence reporting.
Both conditions elicit the same JSON output.

\subsection{Workflow-state detectors}
Three binary indicators are computed from episode text alone by transparent,
auditable regular-expression detectors: $C$ (state conflict: contradiction
patterns and minimization phrases), $M$ (missing information: absence and
unreadability markers), and $H$ (high-risk cues: a red-flag symptom
lexicon). Structural block headers are stripped before detection so template
markers cannot leak the variant label, and the detectors never see gold
labels. Their in-benchmark quality is reported: precision/recall against the
perturbation labels is 0.78/0.73 for $C$, 0.50/0.83 for $M$, and 0.75/0.83
for $H$ (against escalation-required as the target). We additionally probe
lexical dependence with 30 constructed trigger--paraphrase pairs
(10 per detector). Replacements avoid the implemented regex triggers and
span clinical synonyms, register changes, and syntactic rewrites; they were
not sampled from clinical communications or independently adjudicated for
equivalence. Some $C$ pairs probe discourse cues rather than a complete
contradiction. This is an adversarial diagnostic, not a representative
language sample. A frozen \texttt{all-mpnet-base-v2} embedding baseline
assigns each text to its cosine-nearest trigger among the other 29 concepts,
excluding its paired trigger. No paraphrase is a prototype or tuning input.
The artifact pins the model revision, pooling, vectors, and predictions.

\subsection{Workflow-aware selective controller}
The controller post-processes the uncertainty-condition output with a risk
score
\begin{equation}
R = w_u\,(1-\mathrm{conf}) + w_c\,C + w_m\,M + w_h\,H,
\end{equation}
and a two-threshold policy: if $R \ge \tau_{\mathrm{high}}$, override to
\actE{}; if $\tau_{\mathrm{low}} \le R < \tau_{\mathrm{high}}$ and the model
chose \actA{}, override to \actK{}; otherwise keep the model's action.
Weights $w \in \{0,0.25,0.5,0.75,1\}^4$ and thresholds are grid-searched on
the development split only, maximizing the 5:1 utility with decision
accuracy as tiebreaker. No training, gradients, or model internals are
involved; the policy is four multiplications and two comparisons per
episode.

\section{Experimental Setup}
\textbf{Models.} We evaluate five open-weight configurations: Qwen3-1.7B,
Qwen3-4B, Llama-3.2-3B, Gemma-3-4B, and the later-added Qwen3.5-9B local tag
(9.7B parameters). The first four were served with Ollama 0.32.15 on an
Apple M1~Pro (16\,GB); Qwen3.5-9B was served in a separate recorded run with
Ollama 0.24.0 using Q4\_K\_M quantization. We retain separate manifests and
pool predictions only for evaluation. Decoding is deterministic
(temperature 0, fixed seed, 4{,}096-token context, schema-constrained JSON,
thinking disabled for Qwen). The two runs comprise 4{,}000 generations (400
episodes $\times$ 5 models $\times$ 2 conditions) with a 100\% parse rate;
the controller is computed post-hoc.

\textbf{Metrics.} Decision accuracy (action in the acceptable set),
escalation sensitivity/specificity and missed-escalation rate (over episodes
for which the benchmark policy designates \actE{}), cost-sensitive utility at
5:1 and 10:1,
expected calibration error (ECE, 10 bins) and Brier score of confidence
against decision correctness, AUROC of $(1-\mathrm{conf})$ for predicting
decision failure, and risk--coverage curves in which the lowest-scored
episodes are deferred to a human (counted as escalation). We report 95\%
confidence intervals from 5{,}000 bootstrap resamples clustered by semantic
seed. Clean--perturbed differences are averaged within each seed and tested
with paired Wilcoxon signed-rank tests. These choices keep the effective
sample size at 32 independent test seeds rather than treating two surface
parameterizations as independent evidence.

\textbf{Traceability.} Every call is logged with its raw output, latency, and
token counts; the run manifest records model identifiers, model digests,
software versions, and decoding options. The runner is resumable and
seeded; dataset generation is deterministic.

\section{Results}

\begin{table*}[t]
\caption{Main results on the test split ($n=320$; 100 escalation-required).
Sens./Spec.\ are escalation sensitivity/specificity; Missed is the
missed-escalation rate; $U_{5:1}$/$U_{10:1}$ are per-episode cost-sensitive
utilities; AUROC is failure prediction from stated confidence. Controller
rows post-process the uncertainty condition and therefore have no calibrated
probability output (dashes in calibration columns).}
\label{tab:main}
\centering
\begin{tabular}{llccccccccc}
\toprule
Model & Condition & Dec.\ acc. & Sens. & Spec. & Missed & $U_{5:1}$ & $U_{10:1}$ & ECE & Brier & AUROC \\
\midrule
Qwen3-1.7B & vanilla     & 0.388 & 0.02 & 1.000 & 0.98 & $-0.838$ & $-2.369$ & 0.516 & 0.507 & 0.50 \\
           & uncertainty & 0.391 & 0.03 & 1.000 & 0.97 & $-0.819$ & $-2.334$ & 0.544 & 0.539 & 0.48 \\
           & controller  & \textbf{0.694} & \textbf{0.84} & 0.845 & \textbf{0.16} & $\mathbf{+0.488}$ & $+0.238$ & -- & -- & -- \\
\midrule
Qwen3-4B   & vanilla     & 0.556 & 0.46 & 0.986 & 0.54 & $-0.031$ & $-0.875$ & 0.376 & 0.374 & 0.67 \\
           & uncertainty & 0.612 & 0.54 & 0.982 & 0.46 & $+0.112$ & $-0.606$ & 0.305 & 0.305 & 0.77 \\
           & controller  & \textbf{0.825} & \textbf{0.84} & 0.827 & \textbf{0.16} & $\mathbf{+0.462}$ & $+0.212$ & -- & -- & -- \\
\midrule
Qwen3.5-9B & vanilla     & 0.809 & 0.52 & 0.995 & 0.48 & $+0.094$ & $-0.656$ & 0.144 & 0.166 & 0.76 \\
           & uncertainty & \textbf{0.850} & 0.68 & 0.982 & 0.32 & $+0.375$ & $-0.125$ & 0.102 & 0.134 & 0.67 \\
           & controller  & 0.819 & \textbf{0.76} & 0.927 & \textbf{0.24} & $\mathbf{+0.450}$ & $+0.075$ & -- & -- & -- \\
\midrule
Llama-3.2-3B & vanilla     & 0.412 & 0.10 & 1.000 & 0.90 & $-0.688$ & $-2.094$ & 0.448 & 0.424 & 0.72 \\
             & uncertainty & 0.469 & 0.10 & 1.000 & 0.90 & $-0.688$ & $-2.094$ & 0.367 & 0.363 & 0.70 \\
             & controller  & \textbf{0.644} & \textbf{0.84} & 0.845 & \textbf{0.16} & $\mathbf{+0.488}$ & $+0.238$ & -- & -- & -- \\
\midrule
Gemma-3-4B & vanilla     & \textbf{0.775} & 0.64 & 0.982 & 0.36 & $+0.300$ & $-0.262$ & 0.165 & 0.197 & 0.58 \\
           & uncertainty & 0.709 & 0.52 & 0.986 & 0.48 & $+0.081$ & $-0.669$ & 0.161 & 0.212 & 0.71 \\
           & controller  & 0.753 & \textbf{0.88} & 0.832 & \textbf{0.12} & $\mathbf{+0.544}$ & $+0.356$ & -- & -- & -- \\
\bottomrule
\end{tabular}
\end{table*}

\subsection{RQ1: Perturbations expose action-selection failures}
Table~\ref{tab:main} shows the central asymmetry. The Qwen3-1.7B/4B models
reach 0.91 decision accuracy on clean episodes (Table~\ref{tab:variant}), yet
miss 97\%/46\% of required escalations under uncertainty prompting. The
larger Qwen3.5-9B is stronger---0.97 clean accuracy and a 0.32 missed rate---
but is not invariant to state changes: accuracy falls to 0.69 on missing and
0.78 on conflict variants. Across the 20 uncertainty-condition
model--perturbation comparisons, seven paired seed-level Wilcoxon tests
remain significant after Holm correction ($p<0.05$).

\begin{table}[t]
\caption{Decision accuracy by variant (uncertainty condition, test split).}
\label{tab:variant}
\centering
\begin{tabular}{lccccc}
\toprule
Model & Clean & Missing & Stale & Conflict & Late-risk \\
\midrule
Qwen3-1.7B   & 0.91 & 0.09 & 0.91 & 0.00 & 0.05 \\
Qwen3-4B     & 0.91 & 0.28 & 0.94 & 0.12 & 0.81 \\
Qwen3.5-9B   & 0.97 & 0.69 & 0.94 & 0.78 & 0.88 \\
Llama-3.2-3B & 0.70 & 0.44 & 0.61 & 0.41 & 0.19 \\
Gemma-3-4B   & 0.53 & 0.97 & 0.42 & 0.84 & 0.78 \\
\bottomrule
\end{tabular}
\end{table}

\begin{figure}[t]
\centering
\includegraphics[width=\columnwidth]{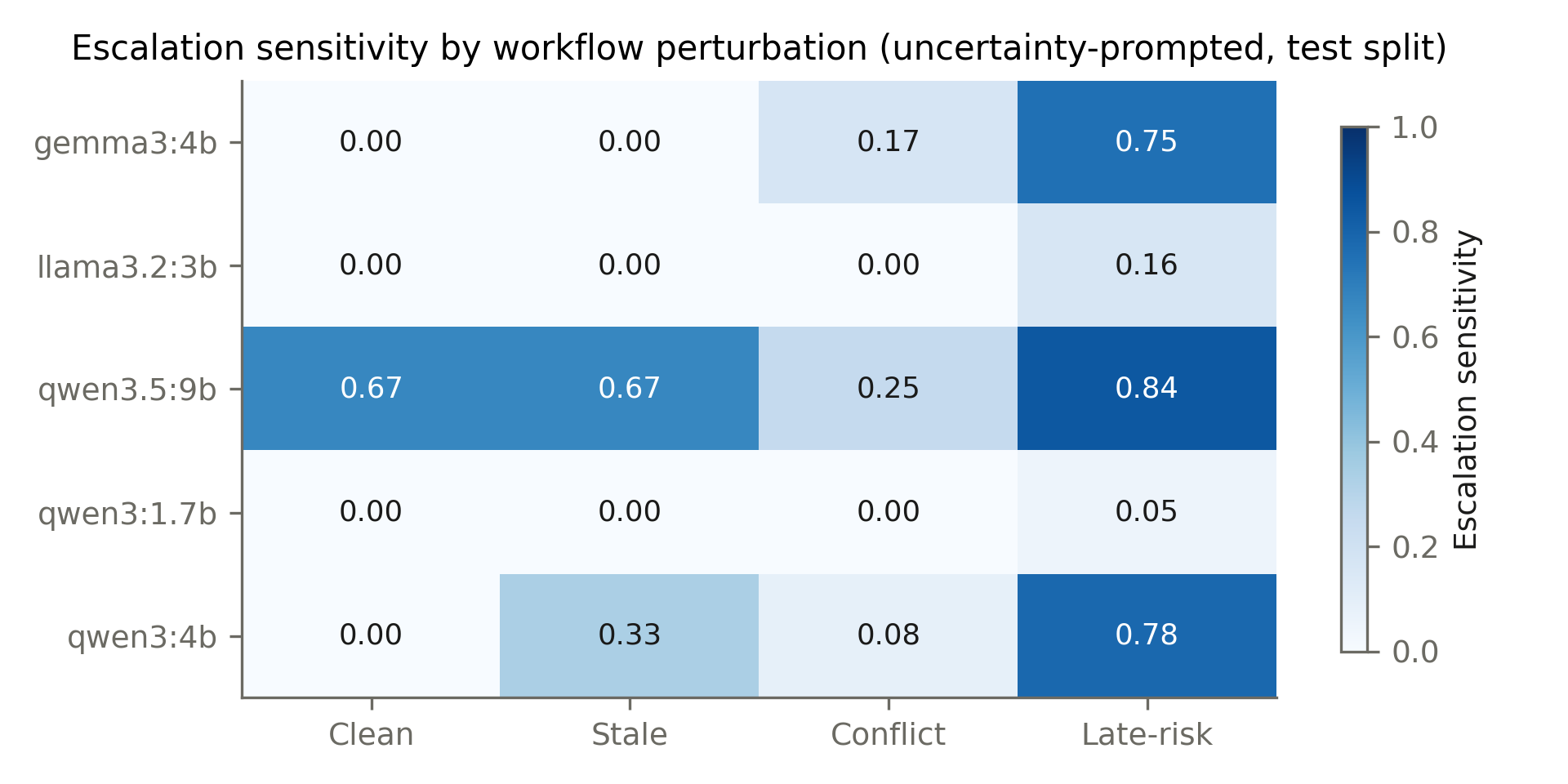}
\caption{Escalation sensitivity by workflow perturbation (uncertainty
condition, test split). The missing variant is omitted: by design it
contains no escalation-required episodes.}
\label{fig:heatmap}
\end{figure}

\begin{figure*}[t]
\centering
\includegraphics[width=0.92\textwidth]{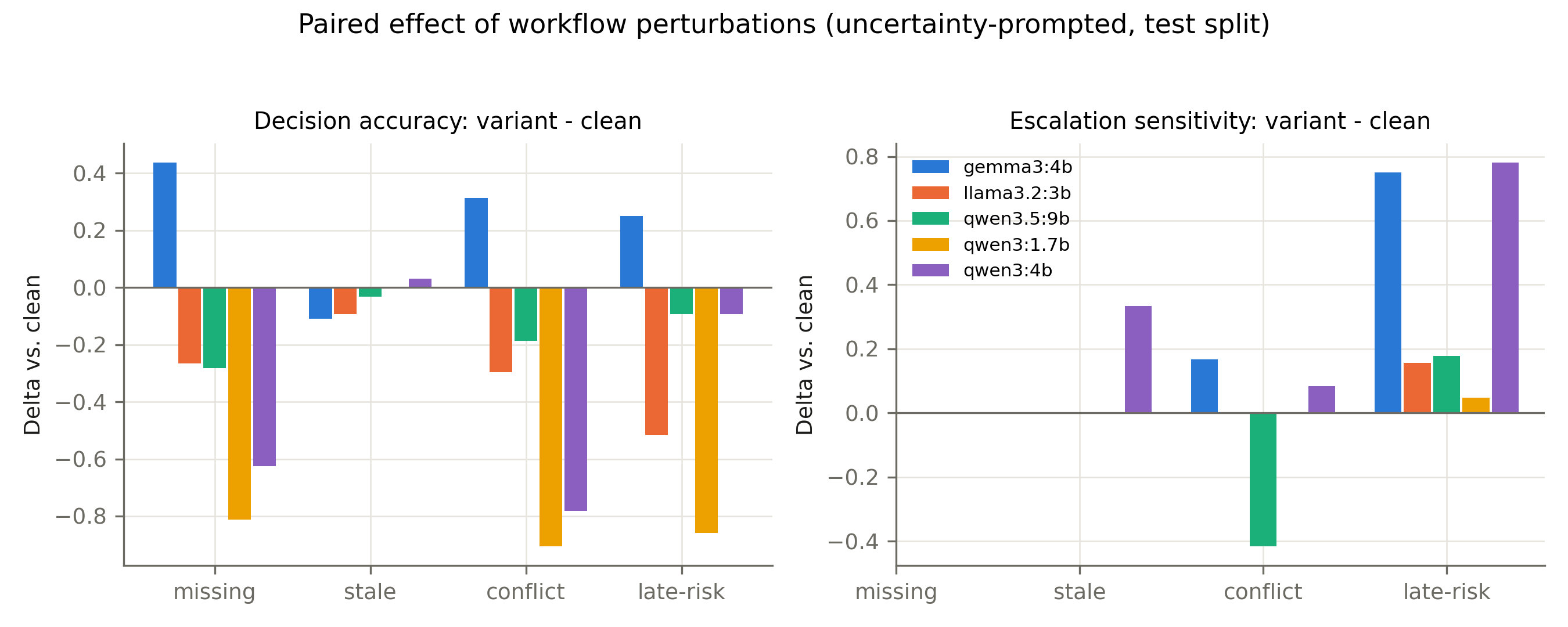}
\caption{Paired effect of each perturbation versus its clean twin
(uncertainty condition, test split). Left: decision accuracy deltas.
Right: escalation sensitivity deltas.}
\label{fig:deltas}
\end{figure*}

Three refinements sharpen the result. First, \textbf{conflict, not
staleness, is the dominant failure axis for the smaller answer-biased Qwen3
models}: both retain 0.91/0.94 accuracy on stale variants but fall to
0.00/0.12 on conflict. Qwen3.5-9B reduces, but does not remove, this gap
(0.94 stale vs.\ 0.78 conflict). Second, \textbf{failures often substitute an
adjacent action} (Table~\ref{tab:actions}): on escalation-required episodes
Llama-3.2-3B asks clarification 70\% of the time while escalating 10\%;
Qwen3-1.7B answers 91\%; Qwen3.5-9B escalates 68\% but asks in another 29\%.
Free-text reasons often name the red flag before under-committing. Third,
\textbf{model action priors interact with the reference shift}: ask-biased
Gemma-3-4B improves on missing and conflict variants, whereas answer-biased
models degrade (Fig.~\ref{fig:deltas}). The late-risk axis separates models:
Qwen3.5-9B, Qwen3-4B, and Gemma-3-4B catch most late red flags
(0.84/0.78/0.75 sensitivity), while Llama-3.2-3B and Qwen3-1.7B rarely do
(0.16/0.05; Fig.~\ref{fig:heatmap}).

\begin{table}[t]
\caption{Action distribution on escalation-required test episodes
(uncertainty condition).}
\label{tab:actions}
\centering
\resizebox{\columnwidth}{!}{%
\begin{tabular}{lcccc}
\toprule
Model & \actB{} & \actA{} & \actK{} & \actE{} \\
\midrule
Qwen3-1.7B   & 0.00 & 0.91 & 0.06 & 0.03 \\
Qwen3-4B     & 0.02 & 0.34 & 0.10 & 0.54 \\
Qwen3.5-9B   & 0.00 & 0.03 & 0.29 & 0.68 \\
Llama-3.2-3B & 0.12 & 0.08 & 0.70 & 0.10 \\
Gemma-3-4B   & 0.04 & 0.00 & 0.44 & 0.52 \\
\bottomrule
\end{tabular}
}
\end{table}

\subsection{Sensitivity to acceptable-action sets}
Set-based accuracy exceeds single-reference exact match by 0--8.8 percentage
points across model/condition pairs (3.8--8.8 for controllers). On the 274
single-action test episodes, including 80 escalation references, controller
5:1 utility improves over uncertainty prompting for four models and ties
for Qwen3.5-9B (both 0.693). The measured gain is sensitive to treatment of
multi-action episodes; complete single-reference and restricted-subset scores are released.

\subsection{RQ2: Verbalized confidence is a weak control signal}
\begin{figure*}[t]
\centering
\includegraphics[width=0.92\textwidth]{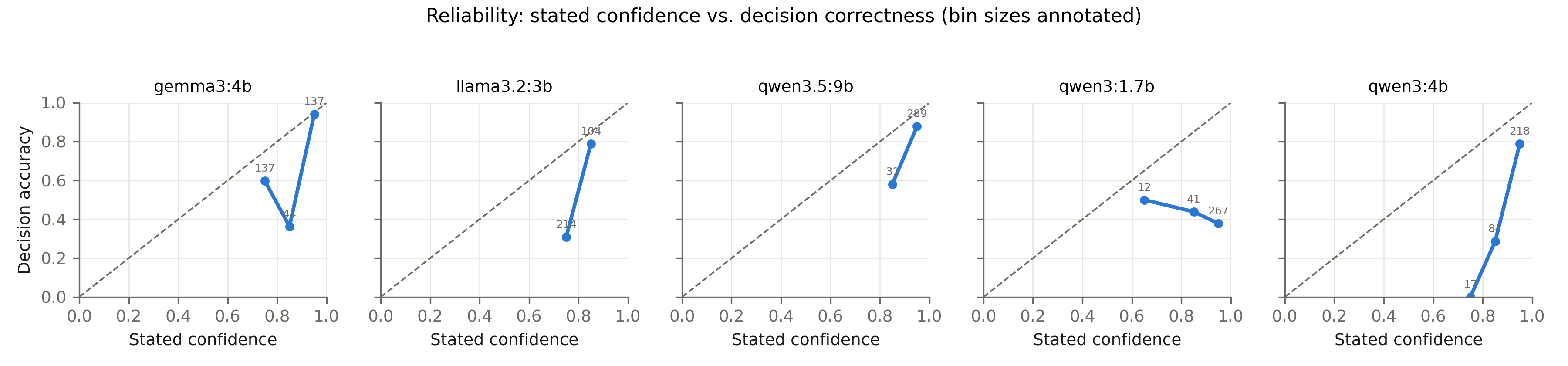}
\caption{Reliability diagrams: stated confidence vs.\ decision correctness
(uncertainty condition, test split; bin sizes annotated). All models are
overconfident, with substantial variation in calibration error.}
\label{fig:reliability}
\end{figure*}

Mean stated confidence is 0.83--0.95 and barely moves
between correct and incorrect decisions. ECE ranges 0.10--0.54; reliability
curves (Fig.~\ref{fig:reliability}) sit far below the diagonal, with most
probability mass in high-confidence bins. AUROC for predicting decision
failure from confidence is chance-level for Qwen3-1.7B (0.48) and
weak-to-moderate for the others (0.67--0.77). Qwen3.5-9B is better calibrated
(ECE 0.10) but its failure-ranking AUROC is only 0.67. More directly, a
confidence-only policy tuned on development data leaves the base action
unchanged for all five models and yields no utility gain. Workflow-only
ablations match the full controller's 5:1 utility for four models and exceed
it for Qwen3.5-9B (0.55 vs.\ 0.45). Thus verbalized confidence can carry a
ranking signal, but it is not sufficient by itself and adds no consistent
value beyond the workflow indicators in this experiment.

\subsection{RQ3: Workflow indicators improve in-benchmark cost-sensitive
routing}
\begin{figure}[t]
\centering
\includegraphics[width=\columnwidth]{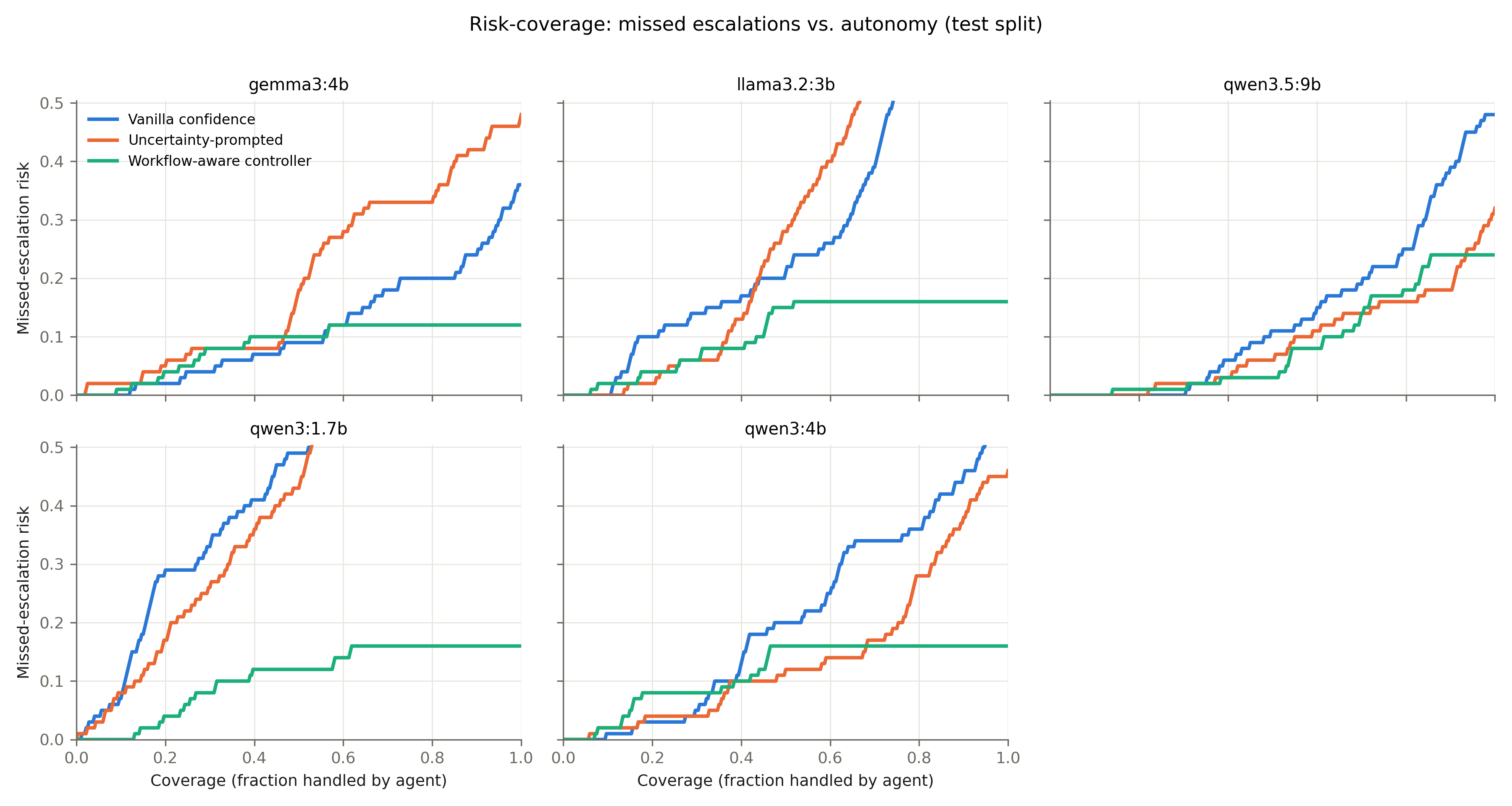}
\caption{Risk--coverage curves (test split): missed-escalation risk as a
function of the fraction of episodes the agent handles itself; deferred
episodes count as escalated. The workflow-aware controller has lower missed
risk than the confidence-based baselines over most moderate-to-high coverage
settings.}
\label{fig:rc}
\end{figure}

The full controller raises escalation sensitivity to 0.76--0.88 and cuts
missed escalations to 0.12--0.24 across the five models, while specificity
falls to 0.83--0.93 (Table~\ref{tab:main}). Under the pre-declared 5:1 cost
matrix, utility improves for every model, from each model's better prompted
baseline ($-0.82\ldots{+0.38}$) to $+0.45\ldots{+0.54}$. It remains positive
at 10:1 ($+0.08\ldots{+0.36}$), while every prompted baseline is negative.
Decision accuracy improves over the uncertainty-prompted action for four of
five models; Qwen3.5-9B instead falls from 0.85 to 0.82, illustrating the
cost-specific nature of the optimization. Risk--coverage curves are
generally lower for the controller at moderate-to-high coverage
(Fig.~\ref{fig:rc}).

The ablations qualify this gain. Workflow-only rules match or exceed full
controller utility for every model, while confidence-only tuning provides no
gain. When thresholds are tuned on three task classes and transferred to the
fourth, 5:1 utility remains nonnegative (0.00--0.90), but the missed rate can
reach 0.50 for evidence retrieval. The supported claim is therefore
in-distribution and cost-conditional: explicit workflow indicators improve
on confidence-only routing here, but neither a universal controller nor a
deployment-ready safety layer has been demonstrated.

\textbf{False escalations.} Of each model's 16--38 nominal false
escalations among 220 non-escalation references, 6--10 permit \actE{} in
the acceptable set. The remaining 6--30 are unaccepted escalations, concentrated
in documentation and patient messaging. Examples include historical or
negated suicidal ideation and ``no blisters'' in a localized rash.
An unbounded regex also matches ``arm\ldots numb'' inside
``pharmacy, lot number.'' These failures expose missing negation, temporal,
and token-boundary handling; they would divert routine work to clinicians
and could produce alert fatigue. Nominal false-positive rates alone
therefore overstate unaccepted actions while understating their varied causes.

\textbf{Broader cost sensitivity.} Table~\ref{tab:costrange} keeps the original
5:1-selected controller fixed. At 1:1, it loses 0.025 utility for Qwen3-4B
and Qwen3.5-9B; at 2:1 the latter ties. All five improve at 3:1 and above,
but all have negative absolute utility at 20:1 ($-0.675$ to $-0.019$).
The artifact includes seed-clustered paired intervals and separately
development-retuned policies; neither analysis establishes a universal
clinical cost ratio.

\begin{table}[t]
\caption{Utility difference versus uncertainty prompting at broader cost
ratios. The controller is fixed at its original 5:1-selected parameters;
these are descriptive post-review analyses.}
\label{tab:costrange}
\centering
\begin{tabular}{lrrrrr}
\toprule
Model & 1:1 & 2:1 & 5:1 & 10:1 & 20:1 \\
\midrule
Qwen3-1.7B & .294 & .547 & 1.306 & 2.572 & 5.103 \\
Qwen3-4B & $-.025$ & .069 & .350 & .819 & 1.756 \\
Qwen3.5-9B & $-.025$ & .000 & .075 & .200 & .450 \\
Llama-3.2-3B & .250 & .481 & 1.175 & 2.331 & 4.644 \\
Gemma-3-4B & .013 & .125 & .463 & 1.025 & 2.150 \\
\bottomrule
\end{tabular}
\end{table}

\begin{figure}[t]
\centering
\includegraphics[width=\columnwidth]{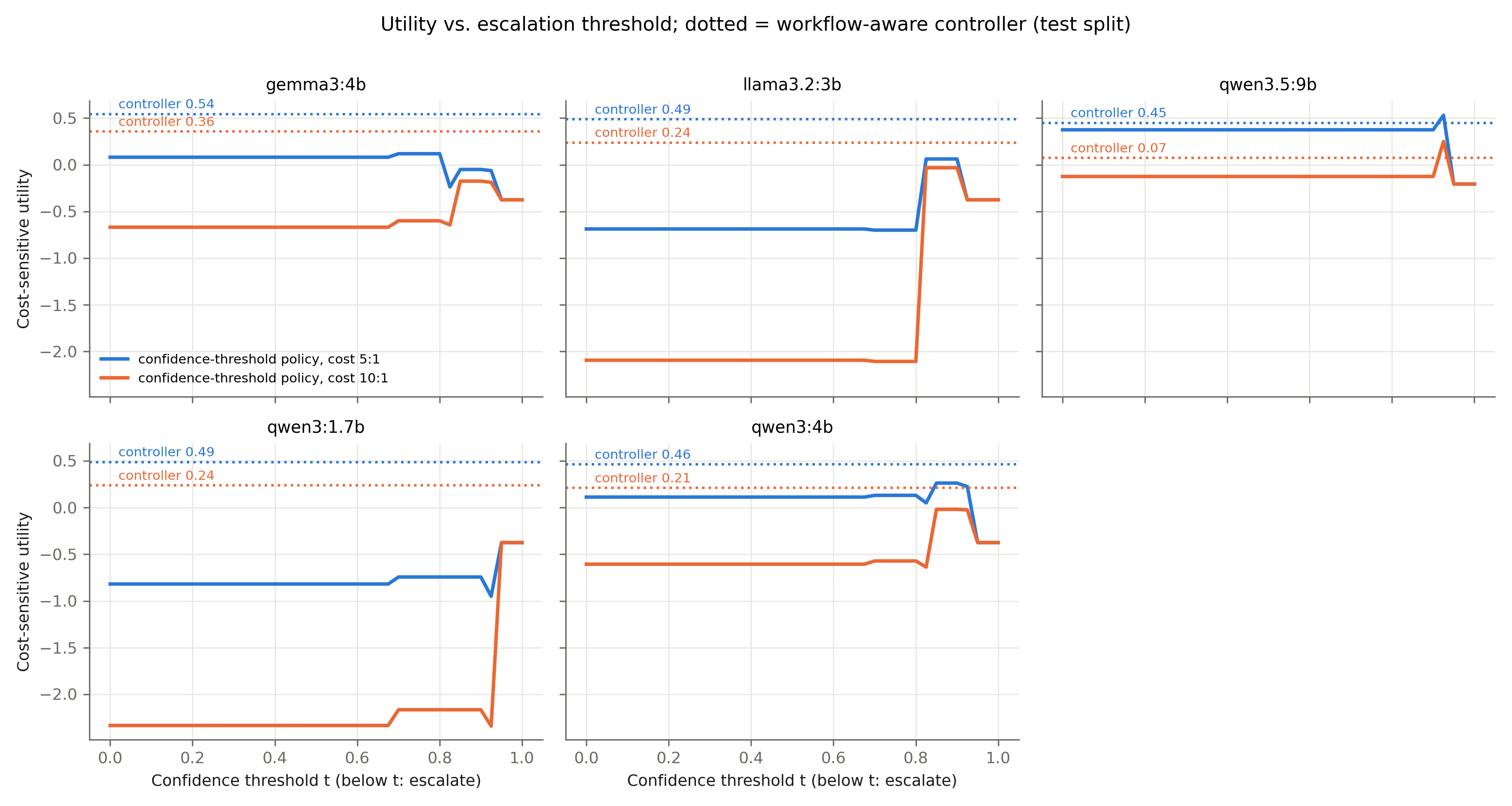}
\caption{Cost-sensitive utility of a confidence-threshold policy (escalate
below $t$) versus the workflow-aware controller (dotted lines), at 5:1 and
10:1 cost ratios (test split).}
\label{fig:utility}
\end{figure}

\subsection{Lexical OOV audit: the controller is not a safety guarantee}
The detector stress set gives a deliberately adversarial mechanism check
(Table~\ref{tab:semantic}). Each regex detector fires on all 10 trigger
examples but none of its 10 paraphrases. Examples include ``shortness of
breath'' $\to$ ``dyspnea,'' ``does not match'' $\to$ ``is inconsistent
with,'' and ``not stated'' $\to$ ``unspecified.'' The frozen embedding
baseline recovers some paraphrases but remains weak on missing-information
cues. Its leave-one-concept-out task assigns one of three cue classes;
the set has no negative/no-cue examples, so neither specificity nor a
deployable multilabel detector is established. We did not reuse cached
model actions to claim end-to-end OOV results: actions could change with
the transcript. The supported conclusion is that the original controller's
gains are in-benchmark and depend on brittle detectors; semantic matching
alone has not established a reliable replacement.

\begin{table}[t]
\caption{Target-cue recall on the 30-concept diagnostic (10 pairs per cue).
Frozen MPNet uses the other 29 trigger texts as prototypes, with each
concept's paired trigger excluded. No clinical or end-to-end validation.}
\label{tab:semantic}
\centering
\begin{tabular}{lcccc}
\toprule
& \multicolumn{2}{c}{Regex} & \multicolumn{2}{c}{MPNet} \\
Cue & Trigger & Paraphrase & Trigger & Paraphrase \\
\midrule
High-risk ($H$) & 1.00 & 0.00 & .90 & .90 \\
Conflict ($C$) & 1.00 & 0.00 & .80 & .60 \\
Missing ($M$) & 1.00 & 0.00 & .40 & .30 \\
\bottomrule
\end{tabular}
\end{table}

\subsection{Uncertainty prompting can backfire}
Explicit uncertainty prompting helped Qwen3-4B (sensitivity $0.46 \to
0.54$, utility $-0.03 \to +0.11$) and Qwen3.5-9B (sensitivity $0.52\to
0.68$, utility $+0.09\to+0.38$), but \emph{hurt} Gemma-3-4B (sensitivity
$0.64 \to 0.52$; clean-variant sensitivity $0.33 \to 0.00$; utility $+0.30
\to +0.08$): the safety checklist pushed the model from \actE{} toward
\actK{}, converting decisive escalations into hedged clarification
requests. Prompt-level safety instructions redistributed probability among
the \emph{cautious} actions rather than toward the \emph{correct} one. This
is a concrete instance of why agent-level behavior must be measured
directly rather than assumed from prompt design.

\subsection{Failure analysis}
High-confidence near-misses illustrate the deficit's character. Faced with
a potassium-repletion request where a newer result of 5.9\,mmol/L (flagged
``hemolyzed?'') contradicted the morning's 3.1, Gemma-3-4B chose \actB{} at
0.95 confidence---correctly identifying the danger in its stated reason,
then declining instead of escalating. Given a levothyroxine refill with
late-arriving palpitations, tremor, and weight loss, it named the
thyrotoxicosis symptoms and still chose \actK{}. Several configurations handled a
classic unstable-angina presentation (exertional chest tightness in a
58-year-old smoker, relieved by rest) as routine scheduling. Qwen3-1.7B,
asked to draft a behavioral-health note after the patient disclosed active
suicidal ideation post-visit, drafted the ``improved'' note anyway at 0.90
confidence. The pattern across failures is consistent: risk recognition
without decision commitment.

\section{Discussion}
The results give the three-level framing of~\cite{yao2026workflow} empirical
content. Models that perform strongly on clean versions of the same workflow
scenarios can still select the wrong action after state perturbations; clean
decision accuracy is not evidence of general medical knowledge. Their
self-reported confidence does not consistently bridge model- and agent-level
behavior. The in-benchmark controller shows that explicit workflow-state
signals can outperform confidence-only routing under the specified
asymmetric costs, but the lexical stress test prevents a deployment claim:
an auditable rule can still be brittle outside the vocabulary used to build
it. The backfiring uncertainty prompt similarly cautions against treating
safety prompting as a validated mitigation; its effect was model-dependent
in both magnitude and sign.

\section{Limitations}
\textbf{Reference labels are single-source and not clinician-validated.} The protocol
of~\cite{yao2026workflow} specifies dual annotation with adjudication and
released disagreements; the present labels were authored by a single
study author. Therefore accuracy is agreement with a study policy, not
clinical correctness. A blinded subset review by two independent clinicians, agreement
analysis, and adjudication would be required to support clinical
interpretation; the present paper limits claims to policy-relative analysis.
\textbf{Small, synthetic, template-assembled
seed pool.} Controlled manipulation is the point---the
paired design follows the precedent of distractor-injection
benchmarks~\cite{vishwanath2025distract}---but surface diversity is
limited. The effective design contains 40 semantic seeds (32 in test), so the
400 episodes are controlled variants, not 400 independent scenarios or a
large-coverage healthcare benchmark. The two parameterizations per seed are
near-duplicates and perturbation phrasing has stylistic regularities a model
could exploit.
\textbf{Detector lexicons were authored with knowledge of the seed pool.}
Controller tuning used only the development split and detectors run on
header-stripped text, but lexical overlap between detector patterns and
episode wording inflates in-benchmark performance; our OOV diagnostic
confirms that the reported controller result is not evidence of lexical or
institutional generalization. Independently authored episodes, natural-language sampling, and
end-to-end semantic-detector evaluation remain needed. \textbf{Open-weight model scope.} One 9.7B model
broadens the original 1.7B--4B evaluation, but no proprietary frontier model
or clinically adapted model is included. \textbf{Over-escalation is real
workload.} The controller's 7--17\% nominal false-escalation rate would
burden clinicians; deployments must set the cost matrix with the governing
clinical team, as the protocol requires, and our conclusions are stated
relative to the pre-declared costs.

\section{Conclusion}
We instantiated a published workflow-aware evaluation protocol as a
400-episode, perturbation-controlled benchmark and audited when five
open-weight LLM configurations escalate. The evaluated systems exhibit
workflow-conditioned action failures that clean-condition accuracy can
conceal; verbalized confidence does not reliably repair them, and
prompt-level safety instructions can worsen them. A transparent controller
improves in-benchmark cost-sensitive utility, while its failure on unseen
lexical realizations shows why mitigation results require their own
generalization tests. The packaged artifact contains episode data, raw model
outputs, manifests, tuned parameters, and evaluation code. Clinician
re-annotation and independent external episodes remain prerequisites for any
clinical claim.

\end{document}